# MULTI-STEP PREDICTION IN LINEARIZED LATENT STATE SPACES FOR REPRESENTATION LEARNING

## A. TYTARENKO

**Abstract.** In this paper, we derive a novel method as a generalization over LCEs such as E2C. The method develops the idea of learning a locallylinear state space, by adding a multi-step prediction, thus allowing for more explicit control over the curvature. We show, that the method outperforms E2C without drastic model changes which come with other works, such as PCC and P3C. We discuss the relation between E2C and the presented method and derived update equations. We provide empirical evidence, which suggests that by considering the multi-step prediction our method - ms-E2C - allows to learn much better latent state spaces in terms of curvature and next state predictability. Finally, we also discuss certain stability challenges we encounter with multi-step predictions and the ways to mitigate them.

## INTRODUCTION

One of the most challenging problems which the field reinforcement learning faces is learning autonomous agents capable of control in Markov decision processes (MDP) with complex state and action spaces. For instance, complactions may arise from large action spaces [1], limited ability to interact with an environment [2], partial observability (POMDP) [3, 4], etc. Optimizing a decent policy takes a lot of samples, usually requires online interactive learning and neural networks capable of processing higher dimensional observations with large number of trainable parameters [5, 6].

There are various algorithms which try to deal with the problem of sample inefficiency, or limited amount of data. Model-based reinforcement learning algorithms [7–9] try to achieve sample efficiency by approximating transition dynamics of an MDP in online or offline mode. Offline reinforcement learning methods [2, 10] strive to extract as much useful information from limited offline data as possible, in order to learn a policy applicable to online regimes as well.

Another algorithmic framework - Learning Controllable Embedding (LCE) - approaches this problem by learning a lower dimensional latent state space and using simpler control algorithms, like iLQR [11], to perform control in this latent space. The challenge here is to make sure that the learned latent space has simpler structure (i.e. next states are easier to predict).

Some particular instances of this framework are described in [9, 12–14]. The idea of E2C [12] is to learn a locally-linear latent space, so that algorithms like LQG could be used for goal-reaching tasks. PCC [13] tries to fix some of the issues encountered in E2C by deriving losses which allow for explicit minimization of latent space's curvature. P3C [14] improves upon PCC mainly by replacing reconstruction loss, needed to make sure the learned state space carries enough information to generate (i.e. decode) observations from latent states. P3C uses predictive coding instead.

In this paper, we seek an alternative approach to enforce lower latent space's curvature and predictability. We generalize E2C by considering multiple transitions at a time, making sure the local linearity is not just preserved between neighbouring states. We inherit the idea of minimization of a joint log likelihood of a transition, generalize it to multiple transitions, and derive a variational bound for further minimization. We then compare the results with LCE approaches and demonstrate a visual representation of learned latent state spaces for a benchmark common among LCE papers.

**PRELIMINARIES**

We denote a **Markov Decision Process** (MDP) $M$ as a tuple $(S, A, r, T)$, where

$S$ - state space,

$A$ - action space

$r : S \times A \to R$ - reward function

$T = P(s_{t+1} | s_t, a_t)$ - probability of state $s_{t+1}$ given current state $s_t$ and action taken $a_t$.

A state of an MDP is a sufficient statistics for a transition kernel, possessing a Markov property.

A task of **Reinforcement Learning** algorithm is for a given MDP $M$ find a policy $\pi$, such that it maximizes the expected return. We are interested in a discounted return objective:

$$\pi^* = \arg\max_\pi E_{\tau \sim p(\cdot)} \sum_{t=0}^{\infty} \gamma^t r(s_t, a_t)$$

Where $\tau$ denotes a trajectory $(x_0, a_0, x_1, a_1, ...)$ obtained by sampling actions using a stochastic policy $\pi$.

In particular, we consider a specific class of Reinforcement Learning algorithms - **Model Based Reinforcement Learning** [7, 8]. Algorithms of this kind usually posses higher sample efficiency, but they involve some sort of an approximation of a transition kernel. LCE algorithms involve parametric models (i.e. neural nets) to learn a good model with desirable properties, like linearity and predictability. This allows to use even the simplest Model-based control (and RL) algorithms like iLQR [11].

**The multi-step Embed to Control model**

Consider an internal transition dynamics of an MDP $M = (S, A, r, T)$:

$$s_{t+1} = f(s_t, a_t) + \omega, \omega \sim P_\omega$$

As we discussed previously, a function $f$ may be highly nonlinear, thus being tricky to optimize with model-based RL or control algorithms. LCE approaches therefore try to learn a mapping from a state space $S$ to some latent space $Z$ such that its latent dynamics

$$z_{t+1} = \hat{f}(z_t, a_t) + \hat{\omega}$$

has some desired properties like local linearity, low curvature, predictability, etc.

In order to learn the mapping $Q_\phi : S \to Z$, Variational Inference framework is employed to derive a tractable algorithm of maximization a likelihood of known data points under the mapping we want to learn.

**Optimization problem**

As follows from the Fig 1, we consider a dataset

$$D = \{(s_t, a_t, s_{t+1}, a_{t+1}, s_{t+2}, ..., a_{t+K-1}, s_{t+K})_i \mid i = 1, ..., N\}$$

containing samples from real trajectories gathered before training. To follow and generalize [12], we define

$$Q_\phi = P_\phi(z_t \mid s_t)$$

as a generative model which for a given state $s_t$ specifies a distribution over the latent space $Z$. Basically, it plays a role of the mapping from $S$ to $Z$ parametrized with a parameter vector $\phi$. And

$$Q_\psi = P_\psi(\hat{z}_{t+1} \mid \hat{z}_t, a_t)$$

as a generative model which for a given latent state $z_t$ and an action $a_t$ predicts the distribution for the next latest state $z_{t+1}$. The model is also parametrized with a parameter vector $\psi$. Also, we denote

$$Q_\psi^j = P_\psi(\hat{z}_{t+j} \mid \hat{z}_{t+i-1}, a_t)$$

In order to find $\phi$ and $\psi$, we maximize the likelihood of a dataset of trajectory samples of length $K$ with respect to the aforementioned parameter vectors:

$$\phi^*, \psi^* = \arg\max_{\phi, \psi} \prod_{i=1}^N P(s_t^i, a_t^i, s_{t+1}^i, a_{t+1}^i, s_{t+2}^i, ..., a_{t+K-1}^i, s_{t+K}^i)$$

For the sake of readability, we denote $s_t, ..., s_{t+K}$ as $s_{t:t+K}$ and $a_t, ..., a_{t+K}$ as $a_{t:t+K}$. Thus our objective is:

$$\phi^*, \psi^* = \arg\max_{\phi, \psi} \prod_{i=1}^N P(s_{t:t+K}^i, a_{t:t+K-1}^i)$$

A corresponding graphical model is depicted on Fig 1.

**Optimization objective**

The objective we defined in a previous subsection is known to be intractable and difficult to optimize. Therefore, LCE approaches employ Variational Inference to find a lower bound to the log-likelihood objective. In this section, we derive this bound for the proposed probabilistic model. **Variational lower bound**:

$$-\log P(s_{t:t+K}, a_{t:t+K-1}) \leq$$

$$E_{\substack{z_t \sim Q_\phi \\ \hat{z}_{t+j} \sim Q_\psi^j \\ j=1,\ldots,K}} \left[ -\sum_{j=1,\ldots,K} \log P(s_{t+j} | \hat{z}_{t+j}) - \log P(s_t | z_t) \right] + D_{KL}[Q_\phi \| P(Z)]$$

Here $D_{KL}[P \| Q]$ denotes Kullback–Leibler divergence functional:

$$D_{KL}[P \| Q] = E_{x \sim P} \log \frac{P(x)}{Q(x)}$$

**Proof**:

$$-\log P(s_{t:t+K}, a_{t:t+K-1}) = -\log \int_{z_t, \hat{z}_{t+1:t+K}} P(s_{t:t+K}, a_{t:t+K-1}, z_t, \hat{z}_{t+1:t+K}) dz_t d\hat{z}_{t+1:t+K}$$

$$= -\log \int_{z_t, \hat{z}_{t+1:t+K}} P(s_{t:t+K} | a_{t:t+K-1}, z_t, \hat{z}_{t+1:t+K}) P(z_t, \hat{z}_{t+1:t+K} | a_{t:t+K-1}) P(a_{t:t+K-1}) dz_t d\hat{z}_{t+1:t+K}$$

$$= -\log \int_{z_t, \hat{z}_{t+1:t+K}} P(s_{t:t+K} | a_{t:t+K-1}, z_t, \hat{z}_{t+1:t+K}) P(z_t, \hat{z}_{t+1:t+K} | a_{t:t+K-1}) P(a_{t:t+K-1}) \frac{Q_\phi}{Q_\phi} dz_t d\hat{z}_{t+1:t+K}$$

$$= -\log \int_{z_t, \hat{z}_{t+1:t+K}} P(s_{t:t+K} | a_{t:t+K-1}, z_t, \hat{z}_{t+1:t+K}) \prod_{j=1}^{K} \left( Q_\psi^j \right) P(z_t) P(a_{t:t+K-1}) \frac{Q_\phi}{Q_\phi} dz_t d\hat{z}_{t+1:t+K}$$

$$= -\log \int_{z_t, \hat{z}_{t+1:t+K}} P(s_t | z_t) \prod_{j=1}^{K} \left( Q_\psi^j P(s_{t+j} | \hat{z}_{t+j}) \right) P(z_t) P(a_{t:t+K-1}) \frac{Q_\phi}{Q_\phi} dz_t d\hat{z}_{t+1:t+K}$$

$$= -\log E_{\substack{z_t \sim Q_\phi \\ \hat{z}_{t+j} \sim Q_\psi^j \\ j=1,\ldots,K}} \left[ P(s_t | z_t) \prod_{j=1}^{K} \left( P(s_{t+j} | \hat{z}_{t+j}) \right) P(a_{t:t+K-1}) \frac{P(z_t)}{Q_\phi} \right]$$

$$\leq E_{\substack{z_t \sim Q_\phi \\ \hat{z}_{t+j} \sim Q_\psi^j \\ j=1,\ldots,K}} \left[ -\sum_{j=1,\ldots,K} \log P(s_{t+j} | \hat{z}_{t+j}) - \log P(s_t | z_t) - \log \frac{P(z_t)}{Q_\phi} \right]$$

$$= E_{\substack{z_t \sim Q_\phi \\ \hat{z}_{t+j} \sim Q_\psi^j \\ j=1,\ldots,K}} \left[ -\sum_{j=1,\ldots,K} \log P(s_{t+j} | \hat{z}_{t+j}) - \log P(s_t | z_t) \right] + D_{KL}[Q_\phi \| P(Z)]$$

**Multi-step embed-to-control model (ms-E2C)**

In this section we instantiate a model for learning a latent locally-linear state space. We use a previously derived upper bound for negative loglikelihood over the multi-step trajectory samples. A graphical models for both E2C and ms-E2C($K$) are shown on the Fig 1.

First, we instantiate **parametric models for encoding and dynamics function** as:

$$Q_\phi = P(z_t | s_t) = N(\mu_\phi(s_t), \Sigma_\phi(s_t)) \text{ - encoder}$$

$$\mu_\phi(s_t), \Sigma_\phi(s_t) = NeuralNet(s_t; \phi)$$

$$Q_\psi^j = P(\hat{z}_{t+j} | \hat{z}_{t+j-1}, a_{t+j-1}) = N(\mu^j{}_\psi, \Sigma^j{}_\psi) \text{ - latent dynamics}$$

$$\mu^j{}_\psi = A_t \mu^{j-1}{}_\psi + B_t a_{t+j-1} + o_t \text{ for } j = 2,\ldots,K$$

$$\mu^j{}_\psi = A_t \mu_\psi + B_t a_t + o_t \text{ for } j = 1$$

$$\Sigma^j{}_\psi = A_t \Sigma^{j-1}{}_\psi A_t^T + \Sigma_W \text{ for } j = 2,\ldots,K$$

$$\Sigma^j{}_\psi = A_t \Sigma_\psi A_t^T + \Sigma_W \text{ for } j = 1$$

$$A_t, B_t, o_t = NeuralNet(z_t; \psi)$$

Thus, given an optimal model would imply a locally linear latent space, in which curvature (i.e. linearity) is explicitly controlled by changing the number of steps per sample. Choosing a large $K$ would recover a globally linear model and setting $K = 1$ recovers an E2C model.

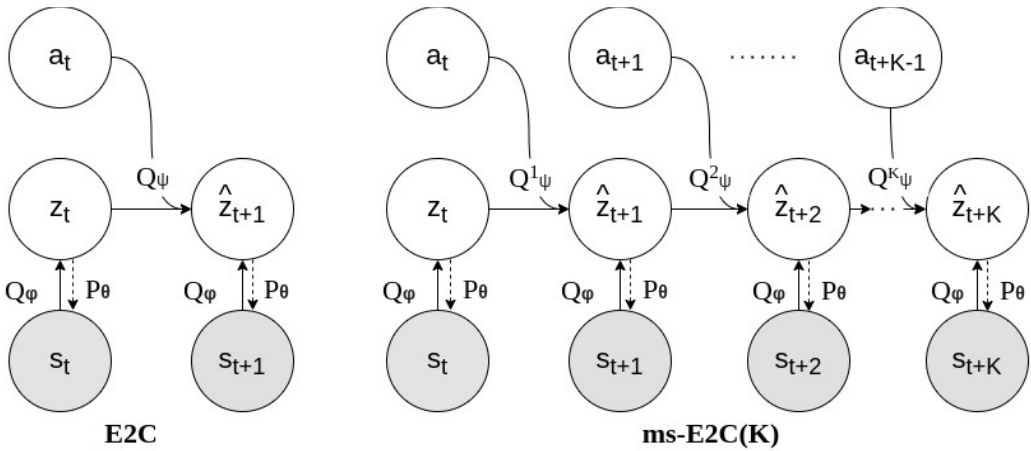

Fig 1. Graphical models for E2C and ms-E2C($K$). Dashed lines denote state reconstruction process. As it follows from the figure, ms-E2C is a generalization of E2C, which one recovers by setting $K = 1$.

We also have to specify a **parametrized decoding model**, which is needed to compute the upper bound, and to enforce a "reconstruction" constraint, introduced in [12] and generalized for our multi-step model:

$$P_\theta = P_\theta(s_t | z_t) = Bernoulli(p(z_t))$$

$$p(z_t) = NeuralNet(z_t; \theta)$$

$$P_\theta^j = P_\theta(s_t \mid \hat{z}_{t+j}) = Bernoulli(p(\hat{z}_{t+j}))$$

$$p(\hat{z}_{t+j}) = NeuralNet(\hat{z}_{t+j}; \theta)$$

**Bernoulli distribution** is chosen for comparability with E2C model on shared benchmark MDPs, i.e. Planar, where the original state space consists of black-and-white images of a grid world with white obstacles and a white circle denoting the position of the agent.

**Loss function.** In order to complete the model's specification, we have to provide a loss function optimizable via stochastic gradient descent. In msE2C it consists of three terms: an estimation of the derived upper bound, consistency term, and stability term.

$$L_{upper}(D_i; \phi, \psi, \theta) = E_{\substack{z_t \sim Q_\phi \\ \hat{z}_{t+j} \sim Q_\psi^j \\ j=1,\ldots,K}} \left[ -\sum_{j=1,\ldots,K} \log P_\theta(s_{t+j} \mid \hat{z}_{t+j}) - \log P_\theta(s_t \mid z_t) \right] + D_{KL}[Q_\phi \parallel P(Z)]$$

The expectation is estimated using a one-sample estimate and a reparametrization trick widely used in variational auto-encoders.

$$L_{consistency}(D_i; \phi, \psi) = \sum_{j=1}^{K} D_{KL}[Q_\psi^j \parallel Q_\phi]$$

$$L_{stability}(D_i; \phi, \psi) = Gersh(A_t(z_t)) + Gersh(B_t(z_t))$$

Here $Gersh(X)$ denotes **Gershgorin loss** [15, 16]:

$$Gersh(X) = \sum_{i=1}^{n} \max(0, X_{i,i} + \sum_{j \neq i} |X_{i,j}| + \varepsilon)$$

where $|X_{i,j}|$ denotes a minor of a matrix $X$, and $\varepsilon > 0$ is a small constant.

According to the Theorem 1 from [15], if the loss value is non-positive, all eigenvalues of a matrix $X$ are guaranteed to have a negative real part, thus ensuring dynamical system stability. The usage of Gershgorin loss in composite loss function is mandatory, as ms-E2C($K$) diverges for larger $K$.

**Algorithm**

Now, we summarize an algorithm for fitting the instance of ms-E2C model we described earlier.

1. Sample a dataset of sub-trajectories using a pretrained or random policy.

$$D = \{(s_t, a_t, s_{t+1}, a_{t+1}, s_{t+2}, \ldots, a_{t+K-1}, s_{t+K})_i \mid i = 1, \ldots, N\}$$

2. Initialize the weights of neural nets $\phi, \psi, \theta$.

3. Repeat for $M$ epochs:

   (a) Retrieve a sample $D_i$ from the dataset $D$

   (b) Compute updated weights using a stochastic gradient descent step:

   $$\phi' = \phi - \gamma \nabla_\phi (L_{upper}(D_i) + \lambda_1 L_{consistency}(D_i) + \lambda_2 L_{stability}(D_i))$$

   $$\psi' = \psi - \gamma \nabla_\psi (L_{upper}(D_i) + \lambda_1 L_{consistency}(D_i) + \lambda_2 L_{stability}(D_i))$$

   $$\theta' = \theta - \gamma \nabla_\theta L_{upper}(D_i)$$

   (c) Update neural networks' parameters:

   $$\phi = \phi'$$
   $$\psi = \psi'$$
   $$\theta = \theta'$$

Here $\lambda_1, \lambda_2$ are tunable hyperparameters.

One might notice that unlike [15] we do not introduce an inner optimization loop to ensure stability of the internal latent space dynamics. Instead, we add the stability loss to the composed loss function. We found that although the difference is apparent during a few first epochs, it becomes negligible after a while. Stability condition does not get violated and the general results are almost the same.

**EXPERIMENTAL VALIDATION**

**Planar system**

Following [12–14], we use a Planar benchmark to compare the performance of the algorithms. In it, a state space is represented as a black-and-white image of a grid world with obstacles. In order to collect a dataset, we sample a random initial state and perform a series of random actions to obtain a trajectory of length $K$.

As in [12, 13], we use a deconvolutional network architecture [17] for image reconstruction from the latent state. For the sake of comparability, we chose the same architecture as in other papers on the topic.

The visualizations of the obtained latent state spaces are provided on a Fig 2. The numerical results are summarized in a Table 1.

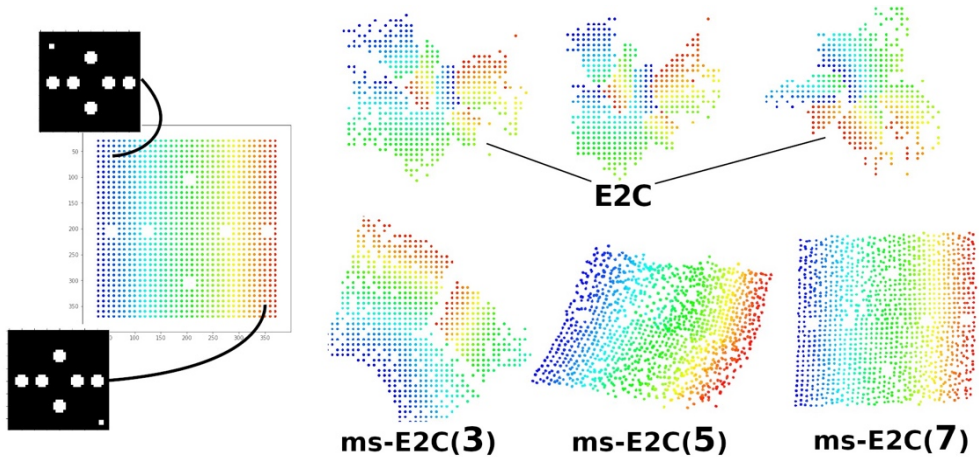

Fig 2. A comparison of latent state spaces learned by E2C and ms-E2C methods. It's worth noting that E2C results coincide with other papers which involved reproduction of E2C [13, 14], while the original paper provides better visuals. A visualization is obtained by transforming all possible environment states with the network $Q_\phi$. See the scheme on the left for details.

Table 1. Comparison of reconstruction and prediction losses. State loss is a regular reconstruction loss. As we observe, ms-E2C($K$) give only slight average improvements on it, which is entirely expected. The introduced method does not change the architecture of a decoding network nor does it add any improvements to the algorithm regarding this matter. An important thing to notice though, is that our generalization does not make the reconstruction performance much worse, which might be expected as representation is influenced by addition prediction constraints. Next state is computed by encoding the state $s_t \xrightarrow{Q_\phi} z_t$, predicting the next latent state $z_t \xrightarrow{Q^1_\psi} z_{t+1}$, and decoding the predicted regular state $z_{t+1} \xrightarrow{P_\theta} s_{t+1}$. Now, the results for previous methods were reproduced with slight perturbations, as we used our own codebase for it.

| Method | State Loss $\log P_\theta(s_t \mid z_t)$ | Next State Loss $\log P(s_{t+1} \mid s_t, a_t)$ |
|---|---|---|
| Non-linear E2C | $9.2 \pm 4.5$ | $11.7 \pm 8.8$ |
| Global E2C | $7.6 \pm 5.7$ | $10.6 \pm 5.2$ |
| E2C | $7.6 \pm 2.3$ | $10.1 \pm 2.7$ |
| ms-E2C(3) | $\underline{7.3 \pm 1.7}$ | $8.7 \pm 1.9$ |
| ms-E2C(5) | $7.6 \pm 2.1$ | $7.5 \pm 1.6$ |
| ms-E2C(7) | $7.7 \pm 2.0$ | $\underline{6.3 \pm 0.9}$ |

# CONCLUSION

In this paper, a novel method had been derived as a generalization over the previous works on LCEs. We demostrate, how the method improves upon E2C without drastic model changes which come with other works, such as PCC and P3C. We empirically show, that by considering the multistep prediction ms-E2C allows to learn a much better latent state spaces in terms of curvature and predictability, by adding a simple yet efficient way to explicitly control the desired curvature of a resulting space. At implementation is available at [18].

Moreover, our work introduces a new dimension to the LCE family of algorithms. Our future work will focus on using the approaches from the state of the art LCE methods, like predictive coding to make LCEs applicable to the higher dimensional real-world MDPs with limited amount of data to learn dynamics embedding from. We will also explore an intriguing possibility to not only encode the state, but also the action space, which sometimes has the complex structure. Lastly, we would like to study various extensions of the method to imitation learning and model-based reinforcement learning.